\documentclass[letterpaper, 10 pt, conference]{ieeeconf}  

\IEEEoverridecommandlockouts                              

\usepackage{graphics} 
\usepackage{epsfig} 
\usepackage{mathptmx} 
\usepackage{times} 
\usepackage{amsmath} 
\usepackage{amssymb}  
\usepackage{algorithm}
\usepackage{cite}
\usepackage{algpseudocode}
\usepackage{multirow}
\usepackage{glossaries}
\usepackage{CJK}
\usepackage{booktabs}
\setacronymstyle{long-short}
\usepackage[dvipsnames]{xcolor}
\usepackage{hyperref}
\usepackage{enumerate}
\usepackage{microtype} 
\DeclareUnicodeCharacter{2212}{-}
\newcommand{\Tau}{\mathrm{T}}

\title{\LARGE \bf A Personalizable Controller for the \\ Walking Assistive omNi-Directional Exo-Robot (WANDER)}

\author{Andrea Fortuna$^{1,2}$, Marta Lorenzini$^{1}$, Mattia Leonori$^{1}$, Juan M. Gandarias$^{1}$\\ Pietro Balatti$^{1}$, Younggeol Cho$^{1}$, Elena De Momi$^{2}$, Arash Ajoudani$^{1}$
\thanks{1- Human-Robot Interfaces and Interaction Laboratory, Istituto Italiano di Tecnologia, Genoa, Italy 2- Department of Electronics, Information and Bioengineering, Politecnico di Milano, Milan, Italy. Corresponding author's email: {\tt\small andrea.fortuna@iit.it}}
\thanks{This work was supported in partby the European Union’s Horizon 2020 research and innovation programme under Grant Agreement No. 871237 (SOPHIA) and by the BRIC LABORIUS Project.}
}

\begin{document}

\maketitle
\thispagestyle{empty}
\pagestyle{empty}

\begin{abstract}
Preserving and encouraging mobility in the elderly and adults with chronic conditions is of paramount importance. 
However, existing walking aids are either inadequate to provide sufficient support to users' stability or too bulky and poorly maneuverable to be used outside hospital environments. In addition, they all lack adaptability to individual requirements.
To address these challenges, this paper introduces WANDER, a novel Walking Assistive omNi-Directional Exo-Robot. 
It consists of an omnidirectional platform and a robust aluminum structure mounted on top of it, which provides partial body weight support. A comfortable and minimally restrictive coupling interface embedded with a force/torque sensor allows to detect users' intentions, which are translated into command velocities by means of a variable admittance controller.
An optimization technique based on users' preferences, i.e., Preference-Based Optimization (PBO) 
guides the choice of the admittance parameters (i.e., virtual mass and damping) to better fit subject-specific needs and characteristics. 
Experiments with twelve healthy subjects exhibited a significant decrease 
in energy consumption and jerk when using WANDER with PBO parameters
as well as improved user performance and comfort.
The great interpersonal variability in the optimized parameters highlights the importance of personalized control settings when walking with an assistive device, aiming to enhance users' comfort and mobility while ensuring reliable physical support. 
\end{abstract}

\section{INTRODUCTION}
\label{sec: introduction}

Gait abnormalities due to illnesses or injuries are one of the main causes of chronic disability in the elderly population \cite{auvinet2017gait}. Their incidence varies greatly with aging as they occur in approximately 35$\%$ of adults aged over 70 years \cite{verghese2006epidemiology} and in 72$\%$ of over 80 \cite{mahlknecht2013prevalence}. 
Consequent balancing and gait difficulties may greatly affect the quality of life \cite{mahlknecht2013prevalence}, restrict the personal independence of those affected, and increase the risk of falls and fall-related injuries \cite{pirker2017gait}. As the proportion of the aging population continues to grow, a corresponding rise in the incidence of gait disorders and related problems is expected, making them a major public health concern. 

To restore, support, and preserve mobility among the elderly, many mechanical structures and appliances for gait assistance have been proposed (e.g. walkers and canes). Nevertheless, the traditional tools suffer from several drawbacks such as requiring sufficient force output to move and handle the device, lack of adaptability with human motion and not providing robust support for user stability.

For this reason, in recent years, robotic-assisted devices have attracted much attention as an alternative. 
Most of them are typically in the form of a robotic cane or smart walker composed of a mobile base and a holding handle  \cite{xing2021admittance, ding2022intelligent, itadera2019predictive, itadera2022admittance, hirata2007development, kikuchi2013evaluation, spenko2006robotic}. Thanks to a combination of sensors and motorized wheels, these systems can better accommodate users' movements and intentions
. However, to be moved, they require users to apply forces on the handlebars, and walking with this bound may result in uncomfortable and non-physiological gait posture. Moreover, they do not prevent vertical falls and, in case of slipping, may not be robust enough to stop the fall. 

These drawbacks have led to an alternative way of gait assistive means i.e., overground walking platforms (OWPs) \cite{alias2017efficacy}. Unlike cane or walker-based systems, an OWP provides partial body weight support (BWS) by ensuring the user to its structure with a system of harnesses thus does not require the use of arms. 
Examples are the KineAssist \cite{peshkin2005kineassist, patton2008kineassist} or the Andago V2.0 \cite{marks2019andago}. The latter, in particular, is a robotic BWS system including a force/torque (FT) sensor coupled to the person that, in combination with the motorized wheels, allows the robotic system to follow the patient.
\begin{figure*}[t]
   \centering
  \includegraphics[trim= 0 1.1cm 0 1.2cm,clip,width=0.82\linewidth]{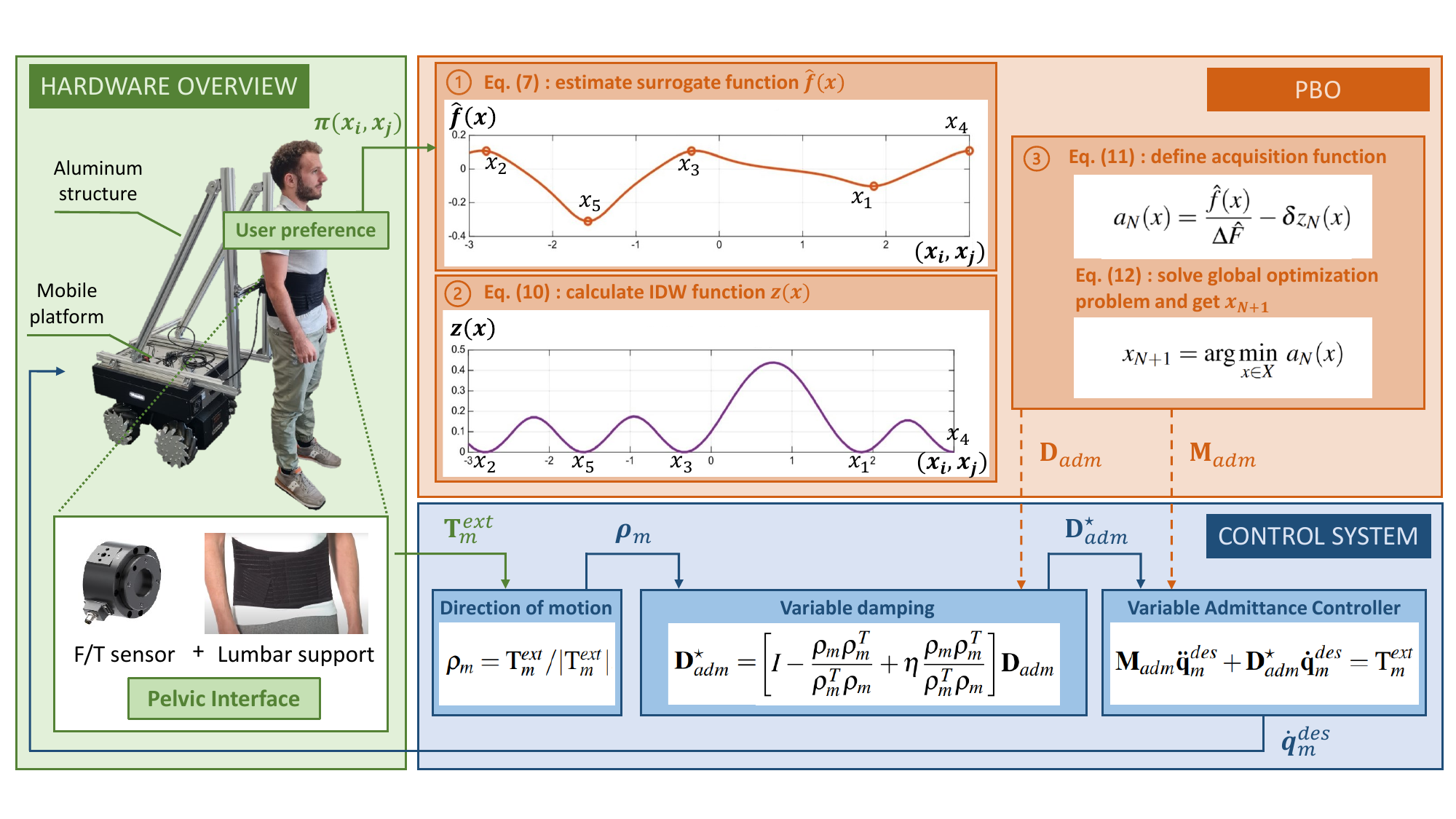}
 \caption{Overview of WANDER and its control framework schema including the preference-based optimization (PBO) of the control parameters.}
\label{fig:Control Framework schema}
\vspace{-0.4cm}
\end{figure*}

In healthcare institutions, this kind of device 
may mitigate the physical effort of nurses and therapists \cite{mun2015development} and empower the workforce, which is limited due to the much higher number of patients with respect to (wrt) medical staff. 
As a result, rehabilitation patients are allowed to increase the training dose and, in general, mobility can be supported for impaired individuals. 
However, since OWPs were designed for hospital environments, they have large dimensions and limited maneuverability, making it difficult to use them elsewhere.

One possibility is to maximize the potential of robotic-assisted devices for usage both in healthcare facilities and at home.
Domestic rehabilitation can complement and continue hospital programs, promoting better recovery in motor capability as it facilitates patients to practice walking and balancing tasks in their home \cite{li2023mobile}. 
On the other hand, many old individuals, in general, 
suffer from a variety of adverse psychosocial difficulties related to falling (e.g., fear, anxiety, loss of confidence, etc.
) that may result in activity avoidance, social isolation and increasing frailty \cite{parry2016cognitive}. Hence, delivering physical support with a home-oriented assistive platform may be highly beneficial for promoting mobility in the elderly. 

With this view, a reduced-dimensions platform named MRBA is developed in \cite{li2023mobile}. 
The proposed user-following algorithm tracks the person’s Center of Mass (CoM) with respect to the robot. If the robot-CoM distance exceeds a predefined threshold, the robot moves toward the user. 
Nevertheless, MRBA platform utilizes non-omnidirectional wheels so the platform may require complex maneuvers for simple movements, which could be uncomfortable for the users. Omnidirectional mobility is highly desirable in OWPs as it removes any restriction on the type of 
motion that can be performed (e.g. walking sideways). 
In this direction, \cite{mun2014design, mun2015development, aguirre2019high, aguirre2021omnidirectional} proposed a novel omnidirectional assistive platform. The latter is provided with an admittance controller that, compared to user-following algorithms, guarantees a smoother human-robot interaction. 
As in the case of the Andago system, this platform needs an FT sensor coupled with the patient for measuring the user’s forces, which are converted into the desired velocity by the admittance model.  
This dynamic relationship 
is highly affected by the choice of the admittance controller parameters. In \cite{mun2014design}, the latter are empirically selected with a trial and error approach, while in \cite{aguirre2019high} and in \cite{aguirre2021omnidirectional}, they are optimized to maximize the sensitivity of the coupled human-robot system. Nevertheless, in all the mentioned works, the parameters remain unchanged for all the users of the platform. 
The type of assistance required by a patient may vary depending on his/her age, weight, and height, the type and severity of the pathology, the level of support the user wants to receive, and the level of stability he/she wants to perceive while using a walking aid. 
This wide range of interpersonal differences makes the establishment of different customized control parameters crucial to cater to the needs of each individual.

This study introduces WANDER, a novel Walking Assistive omNi-Directional Exo-Robot\footnote{An exo-robot is an external stand-alone robot that supports and stabilizes the human body during stances and movements.} with a personalizable controller.
It consists of a mobile omnidirectional platform that ensures free movements on the ground, a robust BWS aluminum structure on top of it, and a comfortable and minimally restrictive coupling interface provided with an FT sensor to detect the users' intentions. The controller includes an admittance interface to map the user-generated forces to the WANDER's base velocity references.     
To personalize the parameters of this controller, thereby enhancing the mobility and comfort of each user while simultaneously preserving the platform's proficiency in providing reliable physical support, we propose an optimization technique named Preference-Based Optimization (PBO). The developed PBO tunes the admittance parameters on-the-fly, based on the subject-specific requirements and preferences while using the system. To evaluate WANDER's effectiveness in reducing energy consumption and jerk while maximizing user performance and comfort against the existing approaches, we conducted experiments with twelve subjects.
The rest of the paper is organized as follows. In Sec.\ref{sec: robotic platform WANDER}, an overview of WANDER and its control framework is provided. Sec. \ref{Sec: Preference based optimization} and Sec. \ref{sec: Experiments} explain the PBO method and the experimental procedure conducted to evaluate it, respectively. In Sec. \ref{sec: results and discussion}, the results are presented and discussed. Sec. \ref{sec: conclusions} draws the conclusions.

\section{Exo-robot for Walking Assistance}
\label{sec: robotic platform WANDER}
In this section, we provide an overview of WANDER and the control system developed to ensure smoothness and transparency with human movements.
\subsection{Hardware Overview}
\label{sec: system overview}
The proposed exo-robot is shown in the green block in Fig. \ref{fig:Control Framework schema}. Its base consists of a velocity-controlled 
Robotnik SUMMIT-XL STEEL mobile platform with three degrees of freedom (DoFs), i.e., able to move in longitudinal and lateral directions and rotate in place. A structure of aluminum profiles is mounted on top of the mobile base and a LaxOne 6-axis FT sensor is assembled in the middle of the horizontal profile. The FT sensor is placed between the structure and the user, who is rigidly coupled to it through a lumbar support at the pelvis level. By means of this pelvic interface (sensor $+$ lumbar support), the forces and torques exerted by the user on the platform can be detected. Its vertical location on the structure can be adjusted to fit the height of each individual.
The range of measurement is equal to $4000 $\;$ N$ in the direction perpendicular to the torso, and $1800  $\;$ N$ along the others (e.g., weight support), while the resolution is about $300  $\;$ mN$ and $400  $\;$ mN$, respectively. The overall dimensions of WANDER are $97.8 \times 77.6 \times 51  $\;$ cm$ and the weight is $150$ kilograms. WANDER is also equipped with a reliable wireless emergency button. 
\subsection{Control System} 
\label{sec: control framework}
The user's intention in terms of force/torques (detected by the FT sensor) can be translated into desired velocities for the platform through an admittance controller (see the blue block in Fig.\ref{fig:Control Framework schema}), making the robot compliant with human motion. 
The underlying dynamic relationship 
can be described by the following admittance model
\begin{equation}\label{Eq: admittance control}
    \mathbf{M}_{adm}\mathbf{\ddot{q}}^{des}_m + \mathbf{D}_{adm}\mathbf{\dot{q}}^{des}_m = \mathbf{\Tau}^{ext}_m,
\end{equation}
where $m = 3$ is the number of DoFs, representing two translational and one rotational movement of the base. $\mathbf{\ddot{q}}^{des}_m$ and $\mathbf{\dot{q}}^{des}_m$ 
are the desired accelerations and velocities, respectively. $\mathbf{\Tau}^{ext}_m = [F_x, F_y, \tau_z]$ are the detected external forces and torques, that are extracted and mapped from the 6D FT sensor data. $\mathbf{M}_{adm}$ and $\mathbf{D}_{adm}$ are the diagonal positive definite matrices of virtual mass and damping in Cartesian coordinates
\[ \mathbf{M}_{adm}=\begin{bmatrix}
M_{x} & 0 & 0 \\
0 & M_{y} & 0 \\
0 & 0 &  J_{z}
\end{bmatrix}, \quad
\mathbf{D}_{adm}=\begin{bmatrix}
D_{x} & 0 & 0 \\
0 & D_{y} & 0 \\
0 & 0 &  D_{z}
\end{bmatrix},
\]
where $M_{x,y}$ and $D_{x,y}$ are mass and damping in the horizontal axes, $J_{z}$ is the moment of inertia and $D_{z}$ is the rotational damping.
In this paper, the robot should be freely moved by the user and maintain its position if no forces are applied. This is why we omit the stiffness term in the admittance model presented in Eq. (\ref{Eq: admittance control}).


To reduce the interaction forces that the users have to apply to move the platform and thus facilitate their movements, less resistance along the direction of motion should be experienced. For this reason, a direction-based variable admittance controller is developed. The motion direction vector is computed according to the FT sensor measurements and expressed as $\rho_m = \mathbf{\Tau}^{ext}_m / \lvert \mathbf{\Tau}^{ext}_m \rvert$. The corresponding projection matrix in the direction of $\rho_m$ is then calculated as in \cite{wang2021variable}
\begin{equation}\label{Eq: projection matrix}
    \mathbf{D}_{mot} = \frac{\rho_m \rho_m^T}{\rho_m^T \rho_m}.
\end{equation}
The initial damping matrix $D_{adm}$ can be decomposed as
\begin{equation} \label{Eq: damping matrix components}
    \left\{\begin{array}{l}
    \mathbf{D}_{\parallel} = \mathbf{D}_{mot} \mathbf{D}_{adm} \\
    \mathbf{D}_{\perp} = I - \mathbf{D}_{mot} \mathbf{D}_{adm}
    \end{array}\right.
\end{equation}
where $ \mathbf{D}_{\parallel}$ and $ \mathbf{D}_{\perp}$ are the parallel and perpendicular components to the direction of motion, respectively.
The damping matrix in Eq. (\ref{Eq: admittance control}) can be thus expressed as
\begin{equation}
    \mathbf{D}_{adm}^{\star} = \mathbf{D}_{\perp} + \eta \mathbf{D}_{\parallel} = \biggl[ I - \frac{\rho_m \rho_m^T}{\rho_m^T \rho_m} + \eta \frac{\rho_m \rho_m^T}{\rho_m^T \rho_m} \biggr] \mathbf{D}_{adm},
\end{equation}
where $\eta \in [0,1]$ is a proportional coefficient to reduce the damping in the direction of motion thus facilitating the users' movements \cite{wang2021variable}.
Eq. (\ref{Eq: admittance control}) is then rewritten accordingly as
\begin{equation}\label{Eq: variable  admittance control}
    \mathbf{M}_{adm}\mathbf{\ddot{q}}^{des}_m + \mathbf{D}^{\star}_{adm}\mathbf{\dot{q}}^{des}_m = \mathbf{\Tau}^{ext}_m.
\end{equation}
Given the above-defined controller, to maximize the users' perception of mobility and comfort while using WANDER, a suitable set of mass and damping parameters must be selected.


\section{Preference-based Optimization}
\label{Sec: Preference based optimization}

In this section, we introduce the PBO method (see the orange block\footnote{The graphs in the orange block representing $\hat{f}(x)$ and $z(x)$ are taken from \cite{bemporad2021global} and illustrate generic functions with an explanatory purpose.} in Fig.\ref{fig:Control Framework schema}) to select the best set of admittance controller parameters (mass and damping) 
based on each user's qualitative feedback while using WANDER.

For some optimization problems, defining a cost function $f$ is not feasible. Instead, it is possible for a user to iteratively express a preference between two different experimental conditions. 
The observed preferences can then be used to iteratively learn a surrogate function $\hat{f}: \mathbb{R}_{n} \to \mathbb{R}$, which tries to approximate the (unknown) function $f$.
This is the principle underlying the global optimization based on active preference learning (GLISp) algorithm proposed in \cite{bemporad2021global}, and also used in \cite{maccarini2022preference, roveda2023human}, which is adopted in this paper.   

Specifically, at each iteration, a set of mass and damping parameters $x_{i} = \{M_{i}, \; D_{i}\} \in \mathbb{R}^{2}$ is proposed to the user, who is asked to test it (by walking with the assistive platform) and compare it to another set $x_j$, expressing a preference $\mathbf{\pi}$ between the two. After $H = N-1$ iterations, 
$N \geq 2$ sets of parameters $x$ 
included in $X = [x_{1}, ..., x_{N}] \in \mathbb{R}^{N\times2}$ are generated with $x_i,x_{j} \in \mathbb{R}^{2}$ such that $x_i \not= x_j \forall i\not= j, i,j = 1,...,N$.

Given two sets of parameters $(x_{i},x_{j})$, the link between preferences $\mathbf{\pi}$ and the objective function $f$ can be simply stated as follows 
\begin{equation}
\label{Eq: preference function}
\mathbf{\pi}(x_{i}, x_{j})=\left\{\begin{split}
-1 & \;\; \text{ if } f(x_{i}) < f(x_{j}) \text{: } x_{i} \text{ preferred to } x_{j},\\
0 & \;\; \text{ if } f(x_{i}) = f(x_{j}) \text{: }  x_{i} \text{ comparable to } x_{j},\\
+1 & \;\; \text{ if }  f(x_{i}) > f(x_{j}) \text{: }  x_{j} \text{ preferred to } x_{i}. 
\end{split}\right.
\end{equation}
The expressed preferences are sequentially added in a preference vector $B = [\mathbf{\pi}(x_{1}, x_{2}), ...,\mathbf{\pi}(x, x^{\star}_N)] \in \{−1,\; 0,\; 1\}^{H} $ until 
the iteration $H_{\text{max}} = N_{\text{max}}-1$ is reached. 
When the user has tested $N$ sets of parameters, the best value up to the corresponding iteration $H$ is denoted as $x^{\star}_N$. As the iterations grow, the algorithm is expected to approach the global optimal set of parameters.

\subsection{Surrogate function}


The surrogate function $\hat{f}$ is parametrized as the following linear combination of Radial Basis Functions (RBFs)
\begin{equation}\label{Eq: surrogate function}
    \hat{f}(x) = \sum_{k=1}^{N}\beta_k\phi(\gamma d(x, x_{k})),
\end{equation}
where $d : \mathbb{R}^{2} \times \mathbb{R}^{2} \to \mathbb{R}$ is the squared Euclidean distance 
\begin{equation*}\label{Eq: euclidean distance}
    d(x, x_{k}) = ||x - x_k ||_{2}^2, \; 
\end{equation*}
$\phi : \mathbb{R} \to \mathbb{R}$ is an RBF, $\gamma > 0$ is a scalar hyper-parameter defining the shape of the RBF,  and $\beta_k$ are coefficients that are determined as explained below. Examples of RBFs are $\phi(\gamma d) = \frac{1}{1+(\gamma d)^2}$ ({inverse quadratic}) and $\phi(\gamma d) = e^{-(\gamma d)^2}$, ({Gaussian}), see more in \cite{bemporad2021global}. 
According to the preference relation in Eq. (\ref{Eq: preference function}),  $\hat{f}$ has to satisfy the following constraints
\begin{equation}
\label{Eq: preference constraints}\begin{split}
    \hat{f}(x_{i}) \leq \hat{f}(x_{j}) - \sigma & \text { if }  \pi(x_{i}, x_{j}) = -1, \\
    \hat{f}(x_{i}) \geq \hat{f}(x_{j}) + \sigma & \text { if }  \pi(x_{i}, x_{j}) = 1,\\
    |\hat{f}(x_{i}) - \hat{f}(x_{j})| \leq \sigma & \text{ if }  \pi(x_{i}, x_{j}) = 0, 
\end{split}    
\end{equation}
where $\sigma > 0$ is a scalar that avoids the trivial solution $\hat{f} \equiv 0$.
Based on the above constraints, the coefficient vector $\beta$ describing the surrogate $\hat{f}$ is obtained by solving the following convex Quadratic Programming (QP) problem:
\begin{flalign}\label{Eq: computation of beta coeff}           
    \min_{\beta, \epsilon} \sum_{h=1}^{H} \varepsilon_{h}+ \frac{\lambda}{2} \sum_{k=1}^{N} \beta_{k}^2 \; \text{ for }
    h \; = \; 1, ..., H\\ \nonumber
    \text{s.t.} \; \; \Lambda \nonumber \leq -\sigma + \varepsilon_h, \qquad \text{ if }  \pi(x_{i(h)}, x_{j(h)}) = -1 \\ \nonumber
    \Lambda \nonumber \geq \sigma - \varepsilon_h, \qquad \text{ if }  \pi(x_{i(h)}, x_{j(h)}) = 1 \\ \nonumber
    |\Lambda| \nonumber \leq \sigma + \varepsilon_h, \qquad \text{ if }  \pi(x_{i(h)}, x_{j(h)}) = 0 \\ \nonumber
    \text{with} \; \; \Lambda = \sum_{k=1}^{N}(\phi(\gamma d (x_{i(h)}, x_k) - \phi(\gamma d(x_{j(h)}, x_k))\beta_k,
\end{flalign}
where $\lambda > 0$ is a scalar that guarantees uniqueness in the solution of the QP problem, $h$ is the iteration index and $\varepsilon_h$ is a positive slack variable that is used to relax the constraint imposed by Eq. (\ref{Eq: preference constraints}) by varying the iteration.\\
The following procedure is then considered: 
\begin{enumerate}[i)]
    \item generate a new sample by pure minimization of the surrogate function $\hat{f}$ defined in Eq. (\ref{Eq: surrogate function})
\begin{equation*}\label{Eq. next point to query}
    x_{N+1} = \text{arg min} \hat{f}(x)\;\; \text{s.t.} \;\; \ell \leq x \leq u
\end{equation*}
with $\beta$ obtained by solving the QP (\ref{Eq: computation of beta coeff}), where $\ell$ and $u \in \mathbb{R}^{2}$ are the lower and upper bounds for $x$, respectively; 
\item ask the user to express a preference $\pi(x_N^{\star}, x_{N+1})$ and update $\hat{f}$ accordingly; 
\item iterate over $N$ until $N_{max}$.
\end{enumerate}
However, purely minimizing the surrogate function (\textit{exploitation}) may lead to converging to a set of optimal parameters $x_N^{\star}$ that is not the global minimum \cite{bemporad2021global}. An \textit{exploration} objective must be also considered to sample other areas of the feasible domain. Such a balance between exploration and exploitation is addressed by defining a proper acquisition function $a : \mathbb{R}^{2} \to \mathbb{R}$, which is minimized instead of the surrogate function $\hat{f}(x)$. The exploration contribution inside the acquisition function is given by the so-called Inverse Distance Weighting (IDW) function defined below.

\subsection{Inverse Distance Weighting function}
To perform the exploration of regions of $\mathbb{R}$ further away from the current best solution in the early iterations and reduce its effect as the number $N$ of tested sets increases, the following IDW function \cite{ZPB22} is used
\begin{eqnarray}\label{Eq: inverse distance weighting variable}
z_N(x)&=& \biggl(1-\frac{N}{N_{max}}\biggr) \tan^{-1} \biggl(\frac{\sum_{k=1}^N w_k(x^{\star}_N)}{\sum_{k=1}^N w_k(x)}\biggr)\\&+& \frac{N}{N_{max}}\tan^{-1}\biggl(\frac{1}{\sum_{k=1}^N w_k(x)}\biggr) \nonumber
\end{eqnarray}
for $x \not\in X$ and $z_N(x)=0$ otherwise, where $w_k(x)=\frac{1}{d(x,x_k)^2}$. 

\subsection{Acquisition function optimization}
Given an exploration parameter $\delta \geq 0$ the acquisition function $a : \mathbb{R}^{2} \to \mathbb{R}$ per set N is then constructed as 
\begin{equation}\label{Eq: acquisition function}
    a_N(x) = \frac{\hat{f}(x)}{\Delta\hat{F}}-\delta z_N(x),
\end{equation}
where $\Delta\hat{F}$ is the range of the surrogate function given $X$
\begin{equation*}\label{Eq: deltaF factor}
    \Delta\hat{F} = \max\limits_{k}\{\hat{f}(x)\} - \min\limits_{k}\{\hat{f}(x)\}
\end{equation*}
that is used in Eq. (\ref{Eq: acquisition function}) as a normalization factor to simplify the choice of the exploration factor $\delta \in [0,1]$.
Given a set of samples $X$ and a preference vector $B$, the next set of parameters $x_{N+1}$ to be tested is computed as the solution of the (non-convex) optimization problem 
\begin{equation}\label{Eq: optimization problem}
    x_{N+1} = \arg\min\limits_{x\in X} \; a_N(x).
\end{equation}
In particular, the Particle Swarm Optimization (PSO) algorithm of \cite{kennedy1995particle} is used to solve Eq. (\ref{Eq: optimization problem}).

\section{Experiments}
\label{sec: Experiments}
This section presents the experiments conducted to validate the PBO method performance in the selection of the optimal admittance parameters for the control of WANDER. 
\subsection{Experimental protocol}
\label{sec: experimental protocol}
Twelve healthy volunteers, six males, and six females, (age: $28.08 \pm 2.31$ years; mass and height in Tab. \ref{Tab:UsersData}), with no history of walking and balance disability, were recruited for the experiments. 
\footnote{The protocol was approved by the ethics committee Azienda Sanitaria Locale (ASL) Genovese N.3 (Protocol IIT{\_}HRII{\_}SOPHIA $554/2020$). Written informed consent was obtained from the participants.}
Prior to the experiments, the pelvic interface was adjusted according to each participant's height. Then, the subjects were connected through it to the platform to obtain a rigid coupling without affecting the users' comfort. 
The experimental procedure consisted of two different phases. 
\subsubsection{Optimization}
In this phase, the optimal admittance controller parameters were obtained for each participant by using the PBO method.
At each iteration, subjects were asked to express a preference between two different sets of mass and damping $(x_i, x_j)$ applied on-the-fly to the controller. To test them, they could move freely in the experimental area, coupled to the platform, without following a predefined path. 
However, they were instructed on the movements that better highlight differences between conditions (e.g., repeating the exact same movements with each set) and encouraged to prove critical situations (e.g., to abruptly start and stop walking). 
\subsubsection{Evaluation}
This phase aimed to verify the PBO method performance. A comparison was made between the subject-specific set of mass and damping parameters obtained by using PBO and two other sets taken from the literature. Specifically, we considered LT1: $[M = 10\; kg, \; D = 120\; Ns/m]$ from \cite{mun2014design, mun2015development} and LT2: $[ M = 33\; kg, \; D = 72,6\; Ns/m]$ from \cite{aguirre2021omnidirectional}.
A within-subjects experiment in which each participant went through all three experimental conditions (i.e., LT1, LT2, and PBO parameters) was performed. 
\begin{figure}[b]
\vspace{-0.5cm}
   \centering
  \includegraphics[trim= 8.5cm 4.8cm 8.5cm 4.5cm,clip,width=\linewidth]{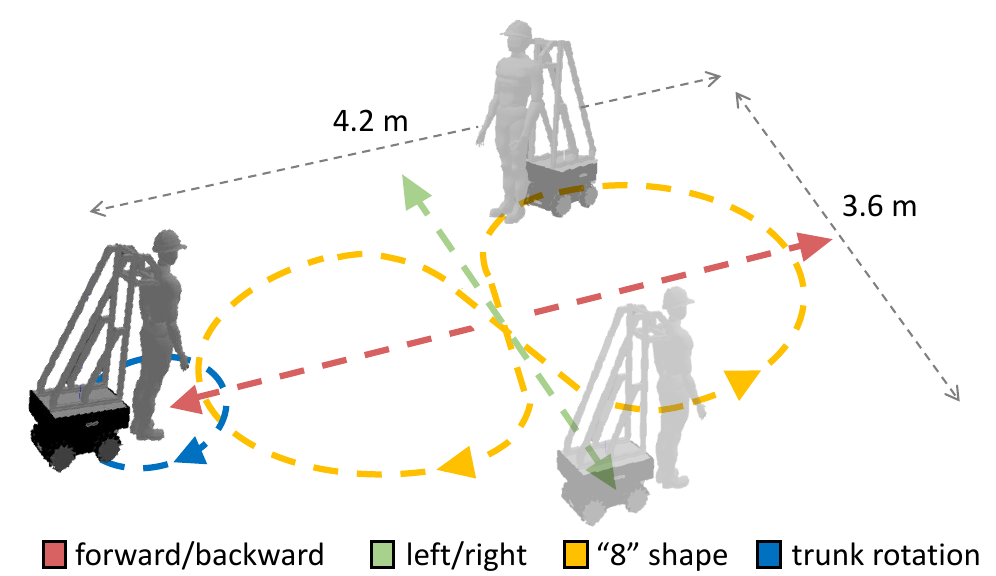}
 \caption{Path followed by the users during the \textit{evaluation} phase.}
\label{fig:Path schema}
\vspace{-0.4cm}
\end{figure}
In each condition, subjects were asked to follow a predefined path, i.e., the tracks represented in Fig. \ref{fig:Path schema}, which were chosen to allow the exploration of all the possible movements that can be performed with the platform.
The specific values selected for the parameters defined in Sec. \ref{sec: robotic platform WANDER} and Sec. \ref{Sec: Preference based optimization} are reported in Tab.\ref{Tab:ExpPar}.
\begin{table}[t]
\centering
\caption{Parameters selected for the controller and PBO.}
\label{Tab:ExpPar}
\begin{tabular}{lll}
\toprule
\textbf{Parameter} & \textbf{Value}     & \textbf{Rationale}             \\
\midrule
$J_{z}$            & $0.33 M_{x,y}$     & Empirical                       \\
$D_{z} $           & $0.33 D_{x,y}$     & Empirical                       \\
$\eta$             & 0.7                & Empirical                       \\
$M_{x,y}$          & $\in [10, \; 100]$ & $9\%$ of system mass\cite{lecours2012variable}, Empirical \\
$D_{x,y}$          & $\in [40, \; 200]$ & Empirical                       \\
$H_{\text{max}}$       & $15$               & Empirical                       \\
RBF function       & Gaussian           & Library default values          \\
$\gamma*$           & $3.0$              & Library default values          \\
$\delta$           & $0.5$              & Library default values          \\
$\sigma$           & $1000$             & Library default values         \\
\bottomrule \\ 
\multicolumn{3}{l}{\footnotesize{$^* \; \gamma$ was recalibrated at $h=9$ to improve the GLISp performance}}
\end{tabular}
\vspace{-5 mm}
\end{table}
To implement the PBO method (Sec. \ref{Sec: Preference based optimization}), we adapted the library proposed in \cite{bemporad2021global,Bem20,ZPB22}. 
\subsection{Experimental analysis}
For the \textit{optimization} phase, the correlation between each subject's weight and the mass obtained with PBO was estimated by using the Pearson coefficient.
For the \textit{evaluation} phase, both quantitative and qualitative metrics were employed and statistical analysis was conducted to test the significance of the results. Specifically, the Wilcoxon signed-rank test was used for pairwise comparisons between the three conditions with a significance level equal to $0.05$.
\subsubsection{Quantitative metrics}
To evaluate the transparency and smoothness of WANDER assistance, respectively, two indicators were defined. The first (for transparency) is the required energy per unit distance (the lower, the better), which comprises the linear energy $E_L$ for forward/backward and lateral movements and angular energy $E_A$ for rotational movements, respectively

\[{E}_{L}= \frac{\int_0^s |F|ds}{s}, \quad
{E}_{A}= \frac{\int_0^\theta |\tau_z|d\theta}{\theta},
\]
where $s$ is the total linear path and $\theta$ is the total angular rotation performed during the task's execution.\\
The second (for smoothness) is the mean value of the module of the jerk
\begin{equation}
    {J}_{mean}= \frac{\sum_{l=0}^{L}\sqrt{J_x^2(l) + J_y^2(l)}}{L},
\end{equation}
where the jerk in the horizontal directions is defined as \cite{wang2018experimental} 
\[J_x = \frac{d^3 q_x}{dt^3}=\frac{d\ddot{q_x}}{dt}, \quad
J_y = \frac{d^3 q_y}{dt^3}=\frac{d\ddot{q_y}}{dt}.
\]
$l$ iterates the acceleration samples collected from the mobile base and $L$ is the total number of samples. 

\subsubsection{Qualitative metrics}
At the end of each experimental condition, participants were asked to fill in the NASA-TLX\cite{HART1988139} to rate perceived workload to vary controller parameters.

\begin{figure*}[t]
    \includegraphics[width=.33\textwidth, height=4.5cm]{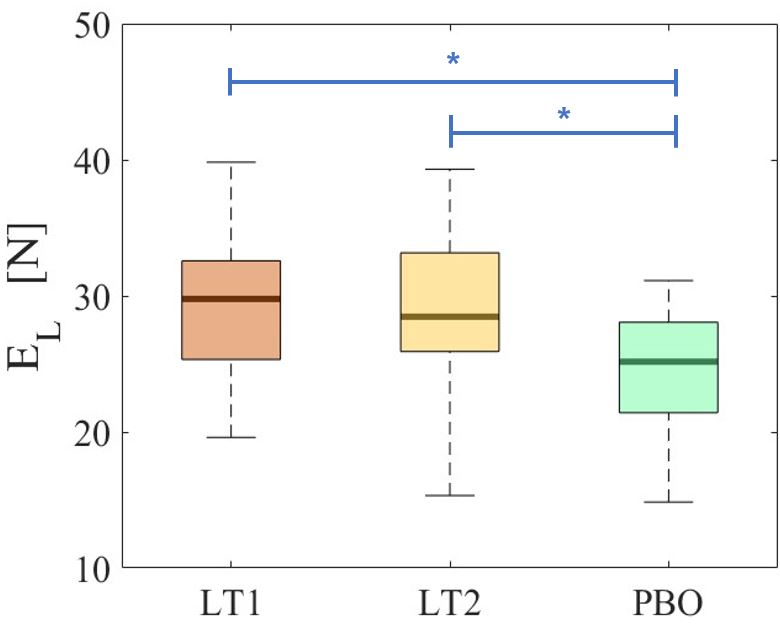}\hfill
    \includegraphics[width=.33\textwidth, height=4.5cm]{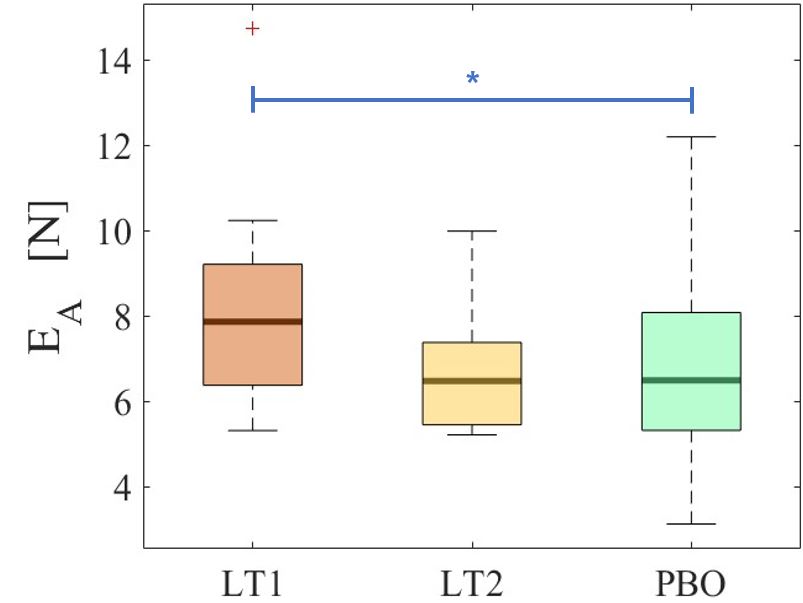}\hfill
    \includegraphics[width=.33\textwidth, height=4.5cm]{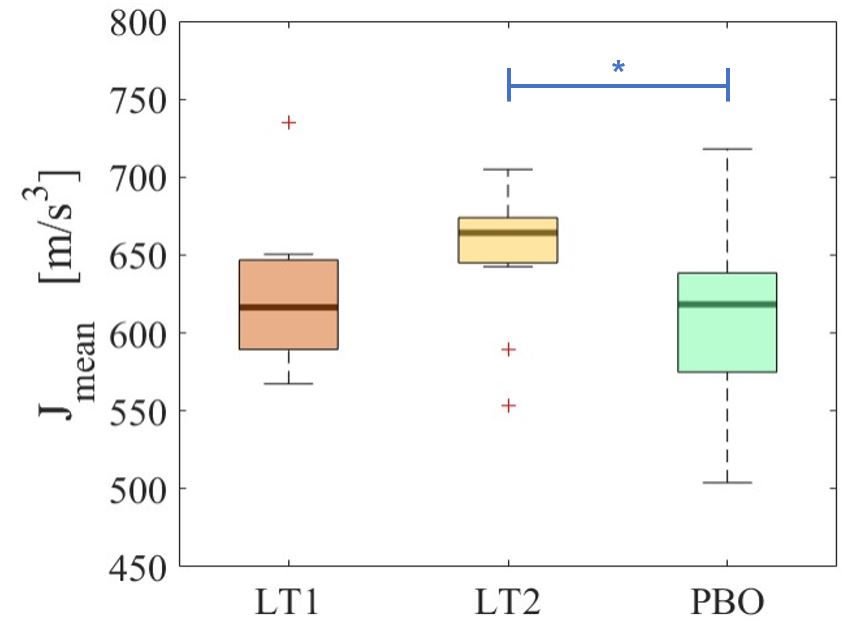}
    \caption{Results of the linear energy, angular energy, and jerk, for all the subjects in the three experimental conditions (LT1, LT2, PBO). By using WANDER with PBO parameters, a jerk profile comparable to a very damped system and a lower energy profile comparable to a lightweight system can be obtained, providing the capabilities of the PBO method in finding optimal parameters. 
    $*$ stands for $p<0.05$.}
    \label{Fig:Res_quant}
    \vspace{-3 mm}
\end{figure*}
\begin{figure}[t]
    \includegraphics[width = \linewidth, trim= 0cm 0 0cm 0,clip]{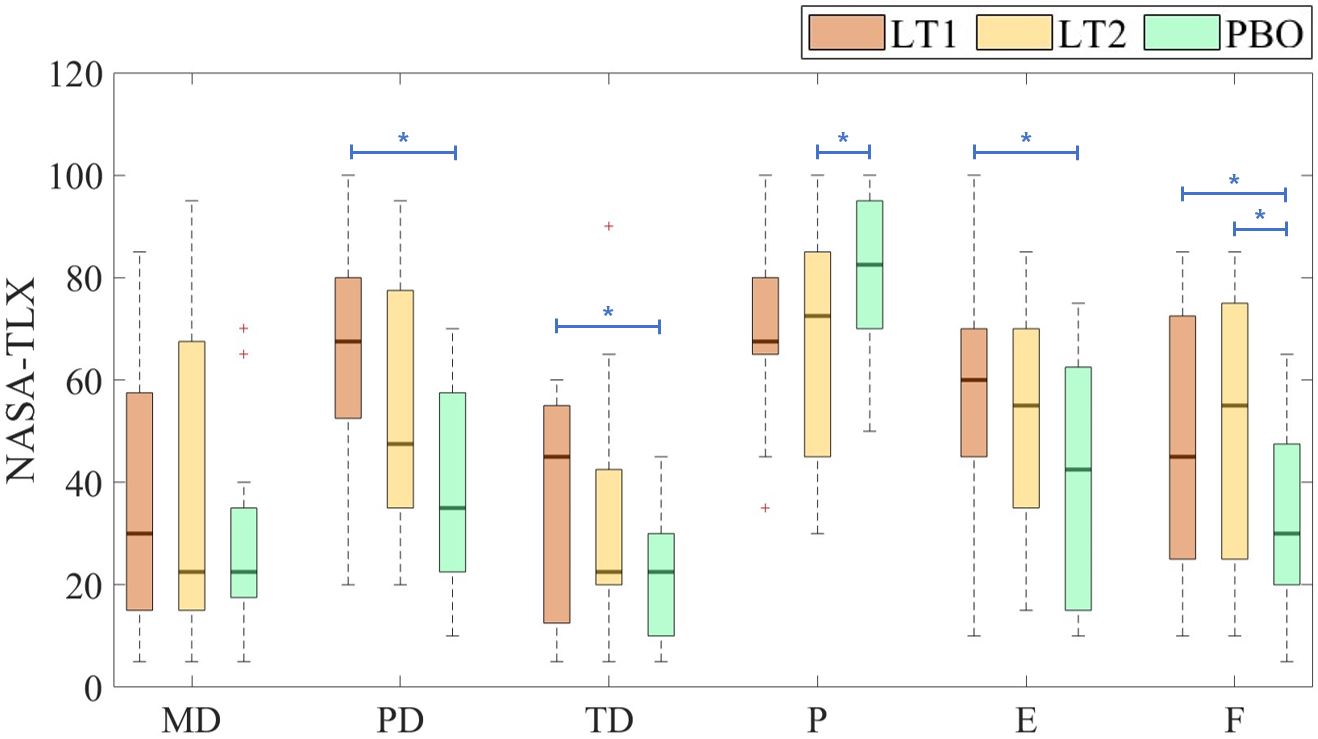}
    \caption{Results of the NASA-TLX for all the subjects in the three experimental conditions (LT1, LT2, PBO). The boxplots for mental demand (MD), physical demand (PD), temporal demand (TD), performance (P), effort (E), and frustration (F) are presented. $*$ stands for $p<0.05$.}
\label{fig:NASA-tlx}
\vspace{-3 mm}
\end{figure}

\section{Results and discussion}
\label{sec: results and discussion}
This section presents and discusses experimental results.
\begin{table}[t]
\caption{Participants' height, weight, and PBO parameters ($M$,$D$).}
\label{Tab:UsersData}
\begin{tabular}{@{}lllll@{}}
\toprule
\textbf{Subject} & \textbf{Height {[}cm{]}}  & \textbf{Weight {[}kg{]}} & \textbf{PBO M {[}Kg{]}}    & \textbf{PBO D {[}Ns/m{]}} \\ \midrule
1       & 180             & 80              & 85.00                & 105.00                    \\
2       & 178             & 70              & 74.30              & 86.00                     \\
3       & 158             & 48              & 81.26             & 40.00                     \\
4       & 183             & 81              & 68.65             & 57.00                   \\
5       & 163             & 61              & 90.00                & 40.00                     \\
6       & 194             & 98              & 100.00               & 40.00                     \\
7       & 155             & 42              & 42.30              & 40.00                     \\
8       & 159             & 65              & 86.95             & 75.80                   \\
9       & 173             & 77              & 68.44             & 57.54                  \\
10      & 176             & 60              & 72.74             & 66.20                   \\
11      & 160             & 49              & 44.80              & 40.00                     \\
12      & 172             & 64              & 89.84             & 62.88                  \\ \bottomrule
\end{tabular}
\vspace{-5 mm}
\end{table}
\subsection{Experimental results}
\label{sec:res:1}
\subsubsection{Quantitative metrics}
In Tab. \ref{Tab:UsersData}, the optimised parameters found with PBO are presented for each subject. A statistically significant correlation equal to $0.59$ (p-value $p = 0.0416$) was found between the users' weights and the masses employed for the admittance controller.
In Fig. \ref{Fig:Res_quant}, the boxplots of the angular energy, linear energy, and jerk, for all the subjects, are presented in the three tested conditions (LT1, LT2, PBO). 
By using the PBO method, the linear energy was significantly decreased on average by $17,61\%$ wrt to LT1 and by $13.93\%$ wrt LT2. 
The angular energy was significantly reduced in PBO wrt LT1 by $13.26\%$ while it remained almost the same in PBO wrt LT2, with a non-significant increase of $3.57\%$. 
The mean jerk was not significantly reduced in PBO wrt LT1 by $1.12\%$ while the decrement of $5.95\%$ between PBO and LT2 was statistically relevant. 
\subsubsection{Qualitative metrics}
In Fig. \ref{fig:NASA-tlx}, the boxplots of the NASA-TLX parameters for all the subjects are reported in all the experimental conditions (LT1, LT2, PBO). 
By using the PBO parameters, the workload perceived by the user was significantly decreased in terms of physical demand, effort, and frustration and increased in terms of performance.

\subsection{Discussion}

As reported in Sec. \ref{sec:res:1}, while the linear energy was significantly reduced by using PBO parameters, the differences among conditions are not relevant for the angular energy. 
This may be due to the higher variability of the rotational movements performed by the users. Indeed, while the linear movements (back/forth, right/left) were easily replicable by all the subjects, rotational movements (as the ``8'' shape path) were often conducted in different ways (e.g., with tighter or wider curves, more or less rotation in place). This suggests the need for more rigorous experimental protocols to test WANDER in different conditions.

Regarding the jerk, it is worth noticing that the LT1 parameters were expected to feature the lowest values, given the highest damping wrt the other conditions.
However, the jerk obtained with PBO parameters is comparable to the one in LT1. 
A positive outcome is also provided by the NASA-TLX, since PBO parameters allow to decrease several indicators.
The fact that, by using WANDER with PBO parameters, we obtained a jerk comparable to a very damped system (LT1), reduced energy, and a milder perceived workload, proves the capabilities of the proposed method in finding optimal parameters.
Hence, we can assert that PBO guarantees an overall better performance wrt to LT1 and LT2 both from a quantitative and qualitative perspective. 

In addition, by analyzing the PBO parameters reported in Tab. \ref{Tab:UsersData}, it is evident that each user requires different controller parameters, which interestingly exhibit a medium-high correlation with the body characteristics. 
This highlights the need for customised and user-specific control settings to better fit individual requirements and demonstrate the benefit of the proposed preference-based approach.

\section{Conclusion}
\label{sec: conclusions}
This work introduced WANDER, an omnidirectional platform with personalized control settings for assisting individuals during walking. Many existing solutions are suitable only for use in care facilities and often lack adaptability to user-specific needs. WANDER has been devised to address these challenges, offering compact dimensions for a more versatile use and a control system that can be tailored to the individual requirements of each user. 
Results prove the effectiveness of user preference-based control parameters in reducing energy and jerk while maximizing user performance and comfort against fixed parameters found in the literature. 


However, this very first version of WANDER presents some limitations. The dynamic interaction between the human and the platform's non-rigid coupling element introduces complexities beyond our current model's scope. Addressing this in future iterations of our work will enhance the system's response.
Also, to enhance the smoothness of human-robot motion, anticipatory direction prediction will be integrated into the variable admittance controller and the inclusion of parameters $J_z$, $D_z$, and $\eta$ in the optimization process will be considered. 
Finally, while this first work focused on enhancing the users' mobility and comfort, the next step will be to use WANDER for balance and gait assistance. 
Additional sensors will be integrated into the platform (e.g., lidar and pressure insoles) to monitor the user's lower limbs. The detection of possible gait anomalies will enable control strategies to recover balance, preventing falls. In this case, the validation will involve the execution of some daily life activities (e.g. sitting on a chair, obstacle avoidance, working on a hob) to prove WANDER portability, dexterity, and applicability in the real world.






\bibliographystyle{IEEEtran}
\bibliography{references.bib} 

\begin{thebibliography}{10}
\providecommand{\url}[1]{#1}
\csname url@samestyle\endcsname
\providecommand{\newblock}{\relax}
\providecommand{\bibinfo}[2]{#2}
\providecommand{\BIBentrySTDinterwordspacing}{\spaceskip=0pt\relax}
\providecommand{\BIBentryALTinterwordstretchfactor}{4}
\providecommand{\BIBentryALTinterwordspacing}{\spaceskip=\fontdimen2\font plus
\BIBentryALTinterwordstretchfactor\fontdimen3\font minus
  \fontdimen4\font\relax}
\providecommand{\BIBforeignlanguage}[2]{{%
\expandafter\ifx\csname l@#1\endcsname\relax
\typeout{** WARNING: IEEEtran.bst: No hyphenation pattern has been}%
\typeout{** loaded for the language `#1'. Using the pattern for}%
\typeout{** the default language instead.}%
\else
\language=\csname l@#1\endcsname
\fi
#2}}
\providecommand{\BIBdecl}{\relax}
\BIBdecl

\bibitem{auvinet2017gait}
B.~Auvinet, C.~Touzard, F.~Montestruc, A.~Delafond, and V.~Goeb, ``Gait
  disorders in the elderly and dual task gait analysis: a new approach for
  identifying motor phenotypes,'' \emph{Journal of neuroengineering and
  rehabilitation}, vol.~14, pp. 1--14, 2017.

\bibitem{verghese2006epidemiology}
J.~Verghese, A.~LeValley, C.~B. Hall, M.~J. Katz, A.~F. Ambrose, and R.~B.
  Lipton, ``Epidemiology of gait disorders in community-residing older
  adults,'' \emph{Journal of the American Geriatrics Society}, vol.~54, no.~2,
  pp. 255--261, 2006.

\bibitem{mahlknecht2013prevalence}
P.~Mahlknecht, S.~Kiechl, B.~R. Bloem, J.~Willeit, C.~Scherfler, A.~Gasperi,
  G.~Rungger, W.~Poewe, and K.~Seppi, ``Prevalence and burden of gait disorders
  in elderly men and women aged 60--97 years: a population-based study,''
  \emph{PloS one}, vol.~8, no.~7, p. e69627, 2013.

\bibitem{pirker2017gait}
W.~Pirker and R.~Katzenschlager, ``Gait disorders in adults and the elderly: A
  clinical guide,'' \emph{Wiener Klinische Wochenschrift}, vol. 129, no. 3-4,
  pp. 81--95, 2017.

\bibitem{xing2021admittance}
H.~Xing, A.~Torabi, L.~Ding, H.~Gao, Z.~Deng, V.~K. Mushahwar, and M.~Tavakoli,
  ``An admittance-controlled wheeled mobile manipulator for mobility
  assistance: Human--robot interaction estimation and redundancy resolution for
  enhanced force exertion ability,'' \emph{Mechatronics}, vol.~74, p. 102497,
  2021.

\bibitem{ding2022intelligent}
L.~Ding, H.~Xing, A.~Torabi, J.~K. Mehr, M.~Sharifi, H.~Gao, V.~K. Mushahwar,
  and M.~Tavakoli, ``Intelligent assistance for older adults via an
  admittance-controlled wheeled mobile manipulator with task-dependent
  end-effectors,'' \emph{Mechatronics}, vol.~85, p. 102821, 2022.

\bibitem{itadera2019predictive}
S.~Itadera, E.~Dean-Leon, J.~Nakanishi, Y.~Hasegawa, and G.~Cheng, ``Predictive
  optimization of assistive force in admittance control-based physical
  interaction for robotic gait assistance,'' \emph{IEEE Robotics and Automation
  Letters}, vol.~4, no.~4, pp. 3609--3616, 2019.

\bibitem{itadera2022admittance}
S.~Itadera and G.~Cheng, ``Admittance model optimization for gait balance
  assistance of a robotic walker: Passive model-based mechanical assessment,''
  in \emph{2022 International Conference on Robotics and Automation
  (ICRA)}.\hskip 1em plus 0.5em minus 0.4em\relax IEEE, 2022, pp. 7014--7020.

\bibitem{hirata2007development}
Y.~Hirata and K.~Kosuge, ``Development of intelligent walker based on passive
  robotics,'' \emph{Seimitsu Kogaku Kaishi/Journal of the Japan Society for
  Precision Engineering}, vol.~73, no.~3, pp. 301--304, 2007.

\bibitem{kikuchi2013evaluation}
T.~Kikuchi, T.~Tanaka, K.~Anzai, S.~Kawakami, M.~Hosaka, and K.~Niino,
  ``Evaluation of line-tracing controller of intelligently controllable
  walker,'' \emph{Advanced Robotics}, vol.~27, no.~7, pp. 493--502, 2013.

\bibitem{spenko2006robotic}
M.~Spenko, H.~Yu, and S.~Dubowsky, ``Robotic personal aids for mobility and
  monitoring for the elderly,'' \emph{IEEE Transactions on Neural Systems and
  Rehabilitation Engineering}, vol.~14, no.~3, pp. 344--351, 2006.

\bibitem{alias2017efficacy}
N.~A. Alias, M.~S. Huq, B.~Ibrahim, and R.~Omar, ``The efficacy of state of the
  art overground gait rehabilitation robotics: a bird's eye view,''
  \emph{Procedia Computer Science}, vol. 105, pp. 365--370, 2017.

\bibitem{peshkin2005kineassist}
M.~Peshkin, D.~A. Brown, J.~J. Santos-Munn{\'e}, A.~Makhlin, E.~Lewis, J.~E.
  Colgate, J.~Patton, and D.~Schwandt, ``Kineassist: A robotic overground gait
  and balance training device,'' in \emph{9th International Conference on
  Rehabilitation Robotics, 2005. ICORR 2005.}\hskip 1em plus 0.5em minus
  0.4em\relax IEEE, 2005, pp. 241--246.

\bibitem{patton2008kineassist}
J.~Patton, D.~A. Brown, M.~Peshkin, J.~J. Santos-Munn{\'e}, A.~Makhlin,
  E.~Lewis, E.~J. Colgate, and D.~Schwandt, ``Kineassist: design and
  development of a robotic overground gait and balance therapy device,''
  \emph{Topics in stroke rehabilitation}, vol.~15, no.~2, pp. 131--139, 2008.

\bibitem{marks2019andago}
D.~Marks, R.~Schweinfurther, A.~Dewor, T.~Huster, L.~P. Paredes, D.~Zutter, and
  J.~C. M{\"o}ller, ``The andago for overground gait training in patients with
  gait disorders after stroke-results from a usability study,''
  \emph{Physiother Res Rep}, vol.~2, no.~2, pp. 1--8, 2019.

\bibitem{mun2015development}
K.-R. Mun, Z.~Guo, and H.~Yu, ``Development and evaluation of a novel
  overground robotic walker for pelvic motion support,'' in \emph{2015 IEEE
  International Conference on Rehabilitation Robotics (ICORR)}.\hskip 1em plus
  0.5em minus 0.4em\relax IEEE, 2015, pp. 95--100.

\bibitem{li2023mobile}
L.~Li, M.~J. Foo, J.~Chen, K.~Y. Tan, J.~Cai, R.~Swaminathan, K.~S.~G. Chua,
  S.~K. Wee, C.~W.~K. Kuah, H.~Zhuo \emph{et~al.}, ``Mobile robotic balance
  assistant (mrba): a gait assistive and fall intervention robot for daily
  living,'' \emph{Journal of NeuroEngineering and Rehabilitation}, vol.~20,
  no.~1, pp. 1--17, 2023.

\bibitem{parry2016cognitive}
S.~W. Parry, C.~Bamford, V.~Deary, T.~L. Finch, J.~Gray, C.~MacDonald,
  P.~McMeekin, N.~J. Sabin, I.~N. Steen, S.~L. Whitney \emph{et~al.},
  ``Cognitive-behavioural therapy-based intervention to reduce fear of falling
  in older people: therapy development and randomised controlled trial-the
  strategies for increasing independence, confidence and energy (stride)
  study.'' \emph{Health Technology Assessment (Winchester, England)}, vol.~20,
  no.~56, pp. 1--206, 2016.

\bibitem{mun2014design}
K.-R. Mun, H.~Yu, C.~Zhu, and M.~S. Cruz, ``Design of a novel robotic
  over-ground walking device for gait rehabilitation,'' in \emph{2014 IEEE 13th
  international workshop on advanced motion control (AMC)}.\hskip 1em plus
  0.5em minus 0.4em\relax IEEE, 2014, pp. 458--463.

\bibitem{aguirre2019high}
G.~Aguirre-Ollinger, A.~Narayan, F.~A. Reyes, H.-J. Cheng, and H.~Yu, ``High
  mobility control of an omnidirectional platform for gait rehabilitation after
  stroke,'' in \emph{2019 IEEE 16th International Conference on Rehabilitation
  Robotics (ICORR)}.\hskip 1em plus 0.5em minus 0.4em\relax IEEE, 2019, pp.
  694--700.

\bibitem{aguirre2021omnidirectional}
G.~Aguirre-Ollinger and H.~Yu, ``Omnidirectional platforms for gait training:
  Admittance-shaping control for enhanced mobility,'' \emph{Journal of
  Intelligent \& Robotic Systems}, vol. 101, pp. 1--17, 2021.

\bibitem{wang2021variable}
Y.~Wang, Y.~Yang, B.~Zhao, X.~Qi, Y.~Hu, B.~Li, L.~Sun, L.~Zhang, and M.~Q.-H.
  Meng, ``Variable admittance control based on trajectory prediction of human
  hand motion for physical human-robot interaction,'' \emph{Applied Sciences},
  vol.~11, no.~12, p. 5651, 2021.

\bibitem{bemporad2021global}
A.~Bemporad and D.~Piga, ``Global optimization based on active preference
  learning with radial basis functions,'' \emph{Machine Learning}, vol. 110,
  pp. 417--448, 2021.

\bibitem{maccarini2022preference}
M.~Maccarini, F.~Pura, D.~Piga, L.~Roveda, L.~Mantovani, and F.~Braghin,
  ``Preference-based optimization of a human-robot collaborative controller,''
  \emph{IFAC-PapersOnLine}, vol.~55, no.~38, pp. 7--12, 2022.

\bibitem{roveda2023human}
L.~Roveda, P.~Veerappan, M.~Maccarini, G.~Bucca, A.~Ajoudani, and D.~Piga, ``A
  human-centric framework for robotic task learning and optimization,''
  \emph{Journal of Manufacturing Systems}, vol.~67, pp. 68--79, 2023.

\bibitem{ZPB22}
M.~Zhu, D.~Piga, and A.~Bemporad, ``{C-GLISp}: Preference-based global
  optimization under unknown constraints with applications to controller
  calibration,'' \emph{IEEE Transactions on Control Systems Technology},
  vol.~30, no.~3, pp. 2176--2187, Sep. 2022.

\bibitem{kennedy1995particle}
J.~Kennedy and R.~Eberhart, ``Particle swarm optimization,'' in
  \emph{Proceedings of ICNN'95-international conference on neural networks},
  vol.~4.\hskip 1em plus 0.5em minus 0.4em\relax IEEE, 1995, pp. 1942--1948.

\bibitem{lecours2012variable}
A.~Lecours, B.~Mayer-St-Onge, and C.~Gosselin, ``Variable admittance control of
  a four-degree-of-freedom intelligent assist device,'' in \emph{2012 IEEE
  international conference on robotics and automation}.\hskip 1em plus 0.5em
  minus 0.4em\relax IEEE, 2012, pp. 3903--3908.

\bibitem{Bem20}
A.~Bemporad, ``Global optimization via inverse distance weighting and radial
  basis functions,'' \emph{Computational Optimization and Applications},
  vol.~77, pp. 571--595, 2020.

\bibitem{wang2018experimental}
C.~Wang, L.~Peng, Z.-G. Hou, L.~Luo, S.~Chen, and W.~Wang, ``Experimental
  validation of minimum-jerk principle in physical human-robot interaction,''
  in \emph{International Conference on Neural Information Processing}.\hskip
  1em plus 0.5em minus 0.4em\relax Springer, 2018, pp. 499--509.

\bibitem{HART1988139}
\BIBentryALTinterwordspacing
S.~G. Hart and L.~E. Staveland, ``Development of nasa-tlx (task load index):
  Results of empirical and theoretical research,'' in \emph{Human Mental
  Workload}, ser. Advances in Psychology, P.~A. Hancock and N.~Meshkati,
  Eds.\hskip 1em plus 0.5em minus 0.4em\relax North-Holland, 1988, vol.~52, pp.
  139--183. [Online]. Available:
  \url{https://www.sciencedirect.com/science/article/pii/S0166411508623869}
\BIBentrySTDinterwordspacing

\end{thebibliography}

\addtolength{\textheight}{-12cm}   

\end{document}